\documentclass[11pt]{article}

\usepackage{a4wide}
\usepackage{amssymb}
\usepackage{amsmath}
\usepackage{amsthm}
\usepackage[dvips]{graphics,color}
\usepackage{epsfig}
\usepackage{natbib}
\usepackage{graphicx}
\usepackage{algorithm2e}

\newcommand{\tr}{^{\text{T}}}
\renewcommand\[{\begin{equation}}
\renewcommand\]{\end{equation}}

\title{Sparse partial least squares for on-line variable selection in multivariate data streams}

\author{ Brian McWilliams \qquad Giovanni Montana \\ Department of Mathematics \\ Imperial College London }

\begin{document}

\maketitle

\begin{abstract}

In this paper we propose a computationally efficient algorithm for on-line variable selection in multivariate regression problems involving high dimensional data streams. The algorithm recursively extracts all the latent factors of a partial least squares solution and selects the most important variables for each factor. This is achieved by means of only one sparse singular value decomposition which can be efficiently updated on-line and in an adaptive fashion. Simulation results based on artificial data streams demonstrate that the algorithm is able to select important variables in dynamic settings where the correlation structure among the observed streams is governed by a few hidden components and the importance of each variable changes over time. We also report on an application of our algorithm to a multivariate version of the "enhanced index tracking" problem using financial data streams. The application consists of performing on-line asset allocation with the objective of overperforming two benchmark indices simultaneously. 

\end{abstract}


\section{Introduction}

Streaming data arise in several application domains, including web analytics, healthcare monitoring and asset management, among others. In all such contexts, large quantities of data are continuously collected, monitored and analyzed over time. Often the main objective is to make real-time predictions by using the incoming streams as covariates in a regression model. In this work, we envisage a system that imports $p$ input and $q$ output data streams at discrete time points. The input data vector is denoted by $x_t \in\mathbb{R}^{1\times p} $ where the subscript refers to the time stamp and the dimension $p$ may be very large. The output $y_t \in\mathbb{R}^{1\times q}$ may also be multivariate. A common task is to recursively estimate a linear regression function of form $y_t=f(x_t)$ which can be used to make future predictions, for instance at time $t+1$. Our fundamental assumption is that, at any given time, only a few selected components of $x_t$ contain enough predictive power, and only those should be actively used to build the regression model. We embrace a penalized regression approach where the unimportant predictors are excluded from the model by forcing their coefficients to be exactly zero.

There are a number of statistical problems arising in this setting which we intend to tackle in this paper. Firstly, a decision has to be made on how to select the truly important predictive components on the input data streams that best explain the multivariate response in a computationally efficient manner. Secondly, since the components of $x_t$ may be highly correlated, variable selection arises in  an ill-posed problem and special care is needed in order to deal with this difficulty. As will be clear later, we take a dimensionality reduction approach. Thirdly, the relationship between input and output streams is expected to change quite frequently over time, with the frequency of change depending on the specific application domain and nature of the data. This aspect requires the development of adaptive methods that are able to deal with possible non-stationarities and the notion of \emph{concept drift}, that is the time-dependency of the underlying data generating process. To the best of our knowledge, little work has been done towards the development of a methodology that resolves all these three issues in a unified framework. 

The problem of tracking latent structures using time varying data streams has been approached in several different ways in the literature. Numerous approaches to on-line principal component analysis (PCA) have been proposed for image analysis \cite{Weng2003} and data stream mining \cite{Papadimitriou2005} amongst others. Tracking and performing regression in the streaming data setting is also well studied with the most well known technique being recursive least squares (see, for example \cite{Haykin2001}). However, the problem of selecting variables on-line has been somewhat less studied. A search of the literature yielded two relatively recent works which address this issue within a penalized regression framework. The earlier method by \cite{Kim2004} proposes a modification to the least angle regression (LARS) algorithm of \cite{LARS} for $L_1$-penalized regression, otherwise known as the Lasso, which allows it to be updated on-line. More recently, \cite{Anagnostopoulos2008} developed an alternative approach to on-line $L_1$-penalized regression based on recursive least squares. The Lasso is solved by using the {\it shooting algorithm}, a pathwise co-ordinate optimization algorithm \cite{Friedman2007}. The resulting procedure is related to adaptive recursive least squares algorithms which have been routinely applied, for instance, in the domain of adaptive filtering. Finally, we note how neither approach considers a multivariate response or the issue of multicollinearity among covariates.

In this work we aim to unify these two problems into a single framework by proposing an efficient incremental and sparse partial least squares (PLS) algorithm for on-line variable selection and tracking of multivariate data streams. PLS regression is an extension of the multiple linear regression model and assumes the existence of a handful of latent factors explaining the variation observed in the data. It has the favorable properties in that it can be used to deal with situations where the data is multicollinear and in problems where the response is multivariate.

The format of this paper is as follows. First, in Section \ref{svd_pls}, we briefly review PLS regression with emphasis on a recent development called Bridge PLS, which was originally proposed for off-line learning by \cite{Gidskehaug2004}. This algorithm is very appealing to us because, unlike other PLS procedures, is not iterative and allows for significant reductions in computational complexity. In Section \ref{spls} we propose a new algorithm to perform sparse Bridge PLS. We achieve sparsification of the regression coefficients by means of a soft-thresholding rule in the computational of the singular value decomposition (SVD). This rule effectively applies a Lasso-like penalty, although many other penalties could be easily used within the same framework. Then in Section \ref{on-line}, our second contribution, an incremental and adaptive version of our sparse Bridge PLS algorithm called incremental Sparse Bridge PLS (iSB-PLS) is proposed for real-time applications. The final algorithm is based on the Adaptive Simultaneous Iterations method for sequential updating of the eigenstructure of a covariance matrix \cite{Erlich1994}. This has the effect of introducing an adaptive behavior, so that changes in the important variables can be tracked in a timely manner. Experimental results using both artificial and real data are presented in Section \ref{results} and conclusive remarks are found in Section \ref{conclusions}. 

\section{Bridge Partial least squares regression} 
\subsection{Partial least squares regression} \label{svd_pls}
Partial least squares (PLS) regression is a method of dimensionality reduction concerned with modeling the relationship between some input data $X\in\mathbb{R}^{n\times p}$  and the response or output $Y\in\mathbb{R}^{n\times q}$ \cite{Hoskuldsson1988}. The assumption underlying PLS is that both $X$ and $Y$ are generated by a small number, $R$, of latent factors
$$
X=\sum_{r=1}^{R}s^{(r)}{b^{(r)}}\tr + E,~~~~
Y=\sum_{r=1}^{R}s^{(r)}{w^{(r)}}\tr + F
$$  
where $s^{(r)}\in\mathbb{R}^{n\times 1}$ are the latent factors and $b^{(r)}\in\mathbb{R}^{p\times1}$ and $w^{(r)}\in\mathbb{R}^{q\times1}$ are the factor loadings of $X$ and $Y$, respectively. $E$ and $F$ are matrices of residuals with no assumed distribution.  PLS finds the latent factors  such that the covariance between input and output is maximized. In order to extract the full complement of latent factors, each one must be extracted sequentially. Once a factor has been extracted, a rank one deflation of the $X$ and $Y$ matrices is performed by subtracting the contribution of the current factor from the data, and a new iteration begins. The PLS literature is extensive and many methods exist for extracting the latent factors (see, for example \cite{Rosipal2006} for a recent review of PLS variants). The various algorithms usually differ beyond computation of the first latent factor by how the input and output data matrices are deflated.

In this work we focus on the commonly used PLS-2 algorithm \cite{Rosipal2006}. The algorithm iteratively finds $R$ hidden factors of $X$ such that $S=XU$ where $S=[s^{(1)},...s^{(R)} ].$  $U=[u^{(1)},..., u^{(R)}]$ is a matrix of weights corresponding to the direction of maximal covariance between $X$ and $Y$. These are found by solving the following optimization problem:
\[
u^{(r)}=\max_{u}[\text{cov}(Xu,Y)]^2  ~~~ s.t.~~ \left\Vert u\right\Vert =1\label{plsr-criterion}
\]
Because it is assumed that $X$ and $Y$ are related through the hidden factors and the factors underlying $X$ are a good predictor of $Y$, the response can be rewritten as 
\[Y=XUW+F \]
This leads to the regression model 
\[
\hat{Y}=X\hat{\beta}+F
\]
where $\hat{\beta}= \tilde{U}\hat{W}$ are the estimated coefficients. For all values of $r$, we define $$M^{(r)}={X^{(r)}}\tr Y^{(r)}$$ that is the covariance matrix between input and output streams. The weight vector $u^{(r)}$ is found by solving Eq \eqref{plsr-criterion} which is equivalent to solving 
\[
\tilde{u}^{(r)}=\arg\max_{u}\left( u\tr M^{(r)}{M^{(r)}}\tr u \right)~~~\text{s.t.}~~\left\Vert u\right\Vert =1 
\]
which is the normalized eigenvector corresponding to the largest eigenvalue of $M^{(r)}{M^{(r)}}\tr $. Alternatively, this is the first left
singular vector of the singular value decomposition (SVD) of $M^{(r)} $. The loading vectors for both $Y$ and $X$ are found by performing univariate regressions 
\[
w^{(r)}=\frac{{s^{(r)}}\tr Y^{(r)}}{{s^{(r)}}\tr s^{(r)}}
 \qquad
 b^{(r)}=\frac{{s^{(r)}}\tr X^{(r)}}{{s^{(r)}}\tr s^{(r)}}
\]
After the extraction of the first factor, in order to extract subsequent factors $X$ and $Y$ must be deflated by subtracting the current latent factor to give
$X^{(r+1)}=X^{(r)}-s^{(r)}{b^{(r)}}\tr $ and
$Y^{(r+1)}=Y^{(r)}-s^{(r)}{w^{(r)}}\tr $. The same procedure is then repeated until all factors are extracted. Clearly, this is not very efficient because it involves the computation of an SVD at each iteration. 

Our first step towards a sparse but also computationally efficient implementation of PLS is to adopt a SVD-based PLS algorithm which extracts the latent factors in a non-iterative way. First, note that the deflation steps above are necessary because if  $\text{rank}(Y)<\text{rank}(X)$, then the covariance matrix $MM\tr $ will be  rank deficient and so the number of PLS components which can be extracted without deflation will be limited to $\text{rank}(Y)$. For instance, in the case of univariate response, $R$ separate SVD computations must be performed. This is the main limiting factor in developing an efficient on-line sparse PLS algorithm that we intent to remove. In order to circumvent this problem, we propose an approach that avoids the deflation steps altogether, thus requiring only one SVD computation for the extraction of all the latent factors.

\subsection{Bridge PLS}
\label{sec-bpls}

Bridge PLS (BPLS) \cite{Gidskehaug2004} is a recent development which ensures that the full complement of PLS components may be extracted in one step by adding a ridge term to the eigenvalue problem. This ensures that the covariance matrix is full rank so we are no longer limited by the rank of Y in the number of components we are able to extract. This is a very important step as it opens the possibility for efficient on-line PLS implementations.

This goal is achieved by introducing a new covariance matrix 
\[
H =\alpha X\tr X + (1-\alpha)MM\tr
\label{bpls}
\]
where $0\leq\alpha\leq1$ is a ridge parameter. It can be noticed that $H$ is a weighted sum between the covariance matrix of $X$ and the covariance matrix of $X$ and $Y$.  When $\alpha=0$, this yields regular PLS and setting $\alpha=1$ yields a principal components regression. Therefore, BPLS can be thought of as biasing the PLS solution towards the PCA solution. The contribution of the ridge parameters can be further seen by rearranging Eq. \eqref{bpls} to obtain
\[
H  = X\tr \left(\alpha I+(1-\alpha)YY\tr \right)X
\label{bpls1}
\]
In this form, it can be noticed that the effect of the ridge parameters is to add a small constant to the diagonal of $YY\tr $. Since 
\[
\text{rank}(\alpha I+(1-\alpha)YY\tr) = \text{rank}(X\tr X)
\]
this prevents $H$ from becoming rank deficient. 

All BPLS weights are then obtained in one step by solving the following modified PLS optimization problem 
\[
\tilde{U}=\arg\max_{U} \left( U\tr H U \right)~~~~\text{s.t.}~\left\Vert U \right\Vert=1
\label{bplscrit}
\]
so that $\tilde{U}=[\tilde{u}^{(1)},..., \tilde{u}^{(R)}] $  are the first $R $ eigenvectors of $H$. The latent factors, $S$ are then computed as $X\tilde{U}$. The corresponding $Y$-loadings are  
\[\hat{W}=\left(S\tr S\right)^{-1}S\tr Y
\]
It is not necessary to compute the $X$-loadings which are normally only required to deflate $X$. The final PLS regression coefficients are given by  $\hat{\beta}= \tilde{U}\hat{W}$. In our experiments we set $\alpha = 10^{-5}$ so that $H$ becomes full rank, yet all PLS directions may be extracted accurately after computing the SVD of $H$ only once; see \cite{Gidskehaug2004} for related discussions.

The computational benefits gained by removing the necessity to perform $R-1$ additional SVD computations in the off-line case is a saving in computation time of $O(Rnp^{2})$. As discussed in the following section, reducing the PLS problem to a single SVD computation provides the key element for performing variable selection in an efficient way in both off-line and on-line scenarios.

\section{New methods for sparse modelling}
 
\subsection{Sparse Bridge PLS \label{spls} } 

In the previous section we briefly reviewed Bridge PLS, a new and efficient method of performing PLS regression which finds the PLS weights by means of a single SVD computation. In this section we observe that the PLS weights can be made sparse by using a penalized form of the SVD which leads us to a novel and efficient method of variable selection based on the Bridge PLS framework.

A regularized SVD method has recently been introduced by \cite{Shen2008} as an efficient device to perform PCA with sparse loading vectors. The method relies on the best low rank approximation property of the SVD. Briefly, this is achieved by reformulating the PCA optimization problem as a regression between $X$ and its best low rank approximation, which is solved by an SVD application. The loading vectors are then made sparse by applying a component-wise thresholding operation.

In this section we use the sparse SVD method of \cite{Shen2008} in order to achieve an efficient variable selection algorithm within the Bridge PLS framework. We first calculate $H$ as in Eq. \eqref{bpls} and define the SVD of  $H=UDV\tr $. The bridge PLS criterion in Eq. \ref{bplscrit} can be written as regression by whereby the criterion to be minimized is the residual sum of squares between $H$ and its low rank approximation, as follows: 
\[
\min_{\tilde{u},\tilde{v}}\left\Vert H-\tilde{u}\tilde{v}\tr \right\Vert ^{2}
\label{Gmin}
\] 
where $\tilde{u}$ and $\tilde{v}\in\mathbb{R}^{p\times1}$ are restricted to be vectors with unit norm so that a unique solution may be obtained. It is known that the product of the first left and right singular vectors, $u^{(1)}v^{(1)}$ is the best rank one approximation of $H$. Therefore Eq. \ref{Gmin} is solved by setting $\tilde{u}=u^{(1)}$ and $\tilde{v}=v^{(1)}$. We obtain sparse loadings by imposing a penalty on $\tilde{u}$ and removing its scale constraint as follows 
\[
\min_{\tilde{u},\tilde{v}}\left\Vert H-\tilde{u}\tilde{v}\tr \right\Vert ^{2}+p(\tilde{u}) ~~~ \text{s.t.} ~~\left\Vert \tilde{v} \right\Vert=1
\label{genpen}
\] 
where $p(\cdot)$ could be one of a number of penalty functions (see, for instance \cite{Friedman2007}). In this work, we concentrate on the Lasso penalty, which places a restriction on the $L_{1}$ norm of $\tilde{u}$. This amount to the following optimization problem:
\[
\min_{\tilde{u},\tilde{v}}\left\Vert H-\tilde{u}\tilde{v}\tr \right\Vert ^{2}+\gamma\left\Vert \tilde{u} \right\Vert
\label{EQspls}
\] 
where $\gamma$ is a parameter which controls the sparsity of the solution. If $\gamma$ is large enough, it will force some variables to be exactly zero. The problem of Eq. \eqref{EQspls} can be solved in an iterative fashion by first setting  $\tilde{u}=u^{(1)}$ and $\tilde{v}=v^{(1)}$ as before. Since $\tilde{u}$ and $\tilde{v}$ are rank one vectors, the Lasso penalty can be applied as a component-wise soft thresholding operation on the elements of $\tilde{u}$ (see, for instance, \cite{Friedman2007}). The sparse $\tilde{u}$ are found by applying the threshold component-wise as follows: 
\begin{eqnarray*}
\label{penalty}
\tilde{u}^{*} & = & \text{sgn}\left(H\tr \tilde{v}\right)\left(\left|H\tr \tilde{v}\right|-\gamma\right)_{+}\\
\tilde{v}^{*} & = & H\tilde{u}^{*}/\left\Vert H\tilde{u}^{*}\right \Vert 
\end{eqnarray*}
We then set $\tilde{u}=\tilde{u}^{*}$ and $\tilde{v}=\tilde{v}^{*}$ and iteratively apply Eq. \eqref{penalty} until $\left \Vert \tilde{u}^{*}-\tilde{u}\right\Vert <\tau $ where $\tau$ is an arbitrarily small constant. The procedure above allows all the PLS weight vectors to be extracted and made sparse at once without the need to  recompute an SVD for each dimension. 

The remaining of the Bridge PLS algorithm proceeds as before, using the newly calculated weights. This leads to latent factors $S=XU$, and the matrix of $Y$ loadings is $W=(S\tr S)^{-1}S\tr Y$. The final sparse PLS regression coefficients are $\hat{\beta}=UW$. Algorithm \ref{ALGsbpls} describes the Sparse Bridge PLS (SB-PLS) procedure in full. 
\vspace{0.5cm}

\begin{algorithm}[H]
\SetKw{KwI}{Initialize}

\KwI{$U=I$, $\gamma = 0$}\;
\KwData{Input $X$ and output $Y$}

\KwResult{Sparse regression coefficients $\beta$}

\SetLine

$M \longleftarrow X \tr  Y$\;
$C \longleftarrow X\tr X$\;
$H \longleftarrow \alpha C+(1-\alpha)MM\tr $\;


$U,D,V\leftarrow \text{SVD}(H)$

\For{$r \leftarrow 1$ \KwTo $R$}{
\While{$\|u^{(r)}-u^{*}\|>\tau$}{

$ \gamma^{(r)} \leftarrow \texttt{findRoot}(u^{(r)})$;


$u^{*} \leftarrow \text{sgn}\left(Hv^{(r)}\right)\left( | Hv^{(r)} | -\gamma^{(r)}\right)_{+}$\;

$v^{(r)}\leftarrow \frac{Hu^{*}}{\left\Vert Hu^{*}\right\Vert}$\;

$u^{(r)} \leftarrow u^{*}$\;

}
$u^{(r)}\leftarrow \frac{u^{(r)}}{\left\Vert u^{(r)}\right\Vert}$\;
} 


$s \leftarrow  x U$\;

$w \leftarrow \frac{ys}{s\tr s}$\;

$\beta \leftarrow Us\tr $\;
\vspace{0.3cm}

\caption{The Sparse Bridge PLS algorithm}
\label{ALGsbpls}
\end{algorithm} 

The parameter $\gamma$ controls the degree of sparsity.  In some situations, such as in financial applications (e.g. Section \ref{sec_finance}), the user may wish to have direct control over the number of variables to be selected. In such a case, it is necessary to select a value of $\gamma$ to induce the correct degree of sparsity in the solution. One naive method of achieving this would be to perform an exhaustive search through the parameter space until a value of $\gamma$ is found which selects the correct number of variables. However, this is inefficient and the value of $\gamma$ which selects the desired number of variables is constantly changing. An alternative consists of using a rootfinding algorithm which performs an efficient search of the parameter space. For instance, we could define a function related to the thresholding operation 
\[
f(\gamma)=\sum_{i=1}^{p}{\mathbb{I} \left(\text{sgn}(u_{i})( | u_{i} | -\gamma)_{+} > 0 \right)} - \theta
\label{pen_func_root}
\]
where $\mathbb{I}$ is an indicator function which finds the non-zero elements of $u$ after the threshold has been applied. Eq \eqref{pen_func_root} performs the componentwise thresholding operation on the weight vector, $u$ and calculates the difference between the number of non-zero elements in $u$ and the target $\theta$, a constant. The rootfinding algorithm is a procedure which finds the value of $\gamma$ such that $f(\gamma)=0$. 

Brent's algorithm is a popular choice as it combines the advantages of other simpler methods (see, for instance, \cite{Forsythe1976}). The most computationally expensive portion of Brent's algorithm is the bisection rootfinding method which is essentially a binary search and so it follows that the maximum additional computational time added is if only the bisection method is applied. The worst case binary search complexity is $O (\log_{2}N)$ where $N$ is the number of possible values that $\gamma$ can take which is determined by the initial guesses $\gamma_1$ and $\gamma_2$. The maximum computational time added by the rootfinding algorithm is $O\left(Rnp\log_{2}N\right)$, i.e. the complexity of the penalization function multiplied by the complexity of the bisection rootfinding algorithm. In practice, some calibration is needed to determine an appropriate initial guess so as to reduce $N$ as much as possible. In our experience, convergence of this specific rootfinding algorithm was normally achieved in less than five iterations. However, our method of choice is a simpler rootfinding algorithm: $\gamma$ is assigned a value equal to the $(p-\theta)^{th}$ largest component of $\left\vert u\right\vert$ where $p$ is the number of elements in $u$. Applying the threshold operation with this value of $\gamma$ will cause all but $\theta$ of the elements in $u$ to become $0$. This replaces the computational effort required to search the parameter space with a much less expensive sort operation of $O(Rp \log p)$ which makes it more suitable for application in an on-line algorithm.

Another sparse PLS algorithm for off-line learning has been proposed by \cite{Le2008}. However, their method is based on the standard PLS regression algorithm described in Section \ref{svd_pls} and thus requires $R$ separate SVD computations to extract all $R$ latent factors. 

\subsection{Incremental Sparse Bridge PLS} \label{on-line}

In this section we develop the Sparse Bridge PLS algorithm to be used for variable selection in the streaming data setting. We call the resulting algorithm incremental Sparse Bridge PLS (iSB-PLS). In this case, we no longer assume we have access to the full data matrix $X\in \mathbb{R}^{n\times p}$. Instead the data arrives sequentially at each time point, $t$, as $x_{t}\in \mathbb{R}^{1\times p}$. Similarly, the response arrives is observable only at discrete time points as $y_{t}\in \mathbb{R}^{1\times q}$.

Although streaming data introduces some challenges, it also offers some computational advantages. For instance, since each observed data vector is of rank one, updating Bridge PLS at each time point is greatly simplified compared to performing Bridge PLS on the full $(n\times p)$ data matrix. The matrix of latent factors is computed as $S=XU$ $\in \mathbb{R}^{1\times R}$. This means the matrix inversion required for the computation of the $Y$-loading matrix reduces to a division by a scalar. 

The main challenge with applying the sparse Bridge PLS algorithm to streaming data is implementing an efficient method to calculate and update the SVD of $H$. Since $H$ is a weighted sum between two covariance matrices we are unable to find its eigenvectors using standard recursive least squares methods. Recursive least squares algorithms require as input the current estimate of the inverse covariance matrix and the new data observation whereas we essentially only have access to a time-varying covariance matrix. Our solution to this problem consists in using the Adaptive SIM algorithm \cite{Erlich1994}, a generalization of the power method which is able to adapt to changes in the data. When a new data point $x_{t}$ and its corresponding response $y_{t}$ arrives, we update the individual covariance matrices as follows
\[
\begin{array}{ccc}
C_{t}&=&\lambda C_{t-1}+x_{t}\tr x_{t} \\
M_{t}&=&\lambda M_{t-1}+x_{t}\tr y_{t}
\end{array}
\]
where $\lambda$ is a forgetting factor which exponentially downweights the contribution of past data points to the current covariance matrix. The Bridge PLS covariance matrix $H_{t}$ of Eq. \eqref{bpls} is constructed by summing the weighted PCA and PLS covariance matrices $C_{t}$ and $M_{t}M_{t}\tr$, which leads to 
\[
H_{t}=\alpha C_{t}+(1-\alpha)M_{t}M_{t}\tr
\]
At each time point, the estimate of the eigenvectors of the covariance matrix, $H$ are updated by performing one iteration of the SIM algorithm as follows:
\[\begin{array}{cc}
Q=&H_{t}U_{t-1} \\
U_{t}=&\text{orth}(Q)
\end{array}
\]
where the function $\text{orth}(Q)$ ensures that the columns of the matrix $Q$ are mutually orthogonal. This allows the columns of $U_{t}$ to converge to different ordered eigenvectors of $H$ as the true eigenvectors of $H$ form an orthogonal basis. This step is necessary becuase, under the power method, every column of $U_{t}$ if left un-normalized will converge to the principal eigenvector of $H$. We use the Gram-Schmidt orthogonalization procedure as follows
\[
\begin{array}{clcl}
u^{(r)}=& \left[ I_{p\times p }-\sum_{k=1}^{r-1}u^{(k)}u^{(k)\text{T}}\right]q^{(r)} ~~~ u^{(1)}=q^{(1)} \\
u^{(r)}=& \frac{u^{(r)}}{\left\Vert u^{(r)} \right\Vert}
\end{array}
\]
which has a computational complexity of $O(pR^2)$.

Once the weight vectors $U_{t}$ have been updated, they are made sparse using the a modified version of the iterative regularized SVD algorithm used for Sparse Bridge PLS in Section \ref{spls}. Since our algorithm is on-line and the solution is updated when a new data point arrives, we no longer iteratively apply the thresholding operation and instead apply it directly to the current estimate of the eigenvector. The simplified sparsification process for the $r^{th}$ weight vector is
\[
\begin{array}{clcl}
u^{*} = & \text{sgn}\left(u^{(r)}\right)\left( | u^{(r)} | -\gamma^{(r)}\right)_{+} \\

u^{*} = & \frac{u^{*}}{\left\Vert u^{*}\right\Vert} 
\end{array}
\]
The final steps of the Bridge PLS algorithm proceed as in the off-line case. The latent vectors $S$ are computed as $S=XU$. However since the number of observations is effectively one, $S$ will be an $R$-vector and the $Y$-loadings can be computed as 
\[
W=Y\tr S/ (S\tr S)
\]
The sparse PLS regression coefficients are $\hat{\beta} = UW$ so that the regression estimate at time $t$ is $\hat{y}_{t}=x_{t}UW$. Algorithm \eqref{ALGisbpls} details the resulting iSB-PLS procedure in full 

\vspace{0.5cm}

\begin{algorithm}[H]
\SetKw{KwI}{Initialize}


\KwI{$U=I$, $m_{0}=0$, $\gamma=0$, $C_{0}=0$}\;
\KwData{Input $x_t$ and output $y_t$} at time $t$

\KwResult{Sparse regression coefficients $\beta_t$} at time $t$

\SetLine






$m_{t} \longleftarrow \lambda m_{t-1}+x_t ^{\text{T}}y_t$\;
$C_{t} \longleftarrow \lambda C_{t-1}+x_t \tr x_t$\;
$H_{t} \longleftarrow \alpha C_{t}+(1-\alpha)m_{t}m_{t}^{\text{T}}$\;

\For{$r\leftarrow 1$ \KwTo $R$}{


$a^{(r)} \leftarrow H_tu^{(r)}$\;
$q^{(r)} \leftarrow \left[ I_{p\times p }-\sum_{k=1}^{r-1}u^{(k)}u^{(k)\text{T}}\right]a^{(r)}$,~~~ $q^{(1)} \leftarrow a^{(1)}$\;
$u^{(r)} \leftarrow q^{(r)}/\left\Vert q^{(r)} \right\Vert$\;
$ \gamma^{(r)} \leftarrow \texttt{findRoot}(u^{(r)})$;


$u^{*} \leftarrow \text{sgn}\left(u^{(r)}\right)\left( | u^{(r)} | -\gamma^{(r)}\right)_{+}$\;

$u^{*}\leftarrow \frac{u^{*}}{\left\Vert u^{*}\right\Vert}$\;

$u_{t}^{(r)} \leftarrow u^{*}$\;

} 


$s \leftarrow  x U_{t}$\;

$w \leftarrow \frac{ys}{s^{\text{T}}s}$\;

$\beta_{t} \leftarrow U_{t}s^{\text{T}}$\;
\vspace{0.3cm}

\caption{The iSB-PLS algorithm}
\label{ALGisbpls}
\end{algorithm}  

In the initialization phase, we set $U_{0}=[u_{0}^{(1)},...u_{0}^{(R)}]=I_{p\times R}$ to ensure that the initial estimates of the eigenvectors are mutually orthogonal. We also initialize $m_{0}=0$, $\gamma=0$, and $C_{0}=0$. The forgetting factor, $\lambda$ is chosen to be between between zero and one. When $\lambda=1$, no data forgetting takes place, whereas $\lambda=0$ has the effect of setting the sample size to the present data point only. Therefore, as the values of $\lambda$ get close to zero, the algorithm becomes more adaptive and the selected variables may change more often. 

In the on-line case the complexity introduced by the penalization function decreases as we operate only on a single data point at a time (i.e. $n=1$). This makes the complexity of the penalization function at each time point $O(Rp)$. 

\section{Experimental results} \label{results}

\subsection{Simulated data}

In this section we report on two simulation experiments designed to demonstrate the performance of the sparse PLS algorithm as an off-line and on-line variable selection method. The input is simulated by first introducing three hidden factors whose temporal evolution is governed by an autoregressive (AR) process of first order in the following way: 
\[
F_{t,j}  =  \delta_j F_{t-1,j}+\epsilon_{t,j} \qquad \text{for } t=2\ldots,400
\]
where $F_{t,j}$ indicates the value of factor $j$ at time $t$, starting with an arbitrary initial value at time $t=1$, and independently for $j=1,2,3$. The parameter $\delta_j$ is the autoregressive coefficient for factor $j$, and we use $\delta_1=0.1$, $\delta_2=0.4$, $\delta_3=0.2$. The error terms in each one of the three factors follow a normal distribution with variance set to $12.25$ and means given by, respectively, $0,-1.5$ and $1.5$. Each input is generated as 
\[
x_{t,i} = F_{t,j} + \eta_{t} \qquad \eta_{t} \sim N(0,1)
\label{datastreams}
\]
where $x_{t,i}$ indicates the values of data stream $i$ at time $t$, for $t=1,\ldots,400$ and $i=1,\ldots,60$. The index $j$ indicates that each stream depends only on a given time-varying hidden factor. Specifically, we create three groups of data streams by setting $j=1$ for $1\leq\text{i}\leq20$, $j=2$ for $21\leq\text{i}\leq40$ and $j=3$ for $41\leq\text{i}\leq60$. 

Using these simulated data streams, we show that the off-line sparse Bridge PLS can accurately select the correct variables where the underlying factors which make up the response do not change over time. We also show how, for such stationary data, both on-line and off-line algorithms lead to the same solution after convergence has taken place in the on-line case. In the off-line case, we consider only the first 100 data points and create a univariate response variable by assigning coefficients to one group of variables, which have been sampled from a normal distribution centered at $10$ and with low variance. Likewise, we assign smaller valued coefficients to the second group of variables by sampling from a normal distribution centered at $5$ with a low variance. The third group of variables are designated "inactive variables" and assigned a zero coefficient. For ease of visualization and interpretation of the results, we have chosen to define a univariate response, however the SB-PLS and iSB-PLS algorithms can also be used in cases where the response is multivariate (e.g. see Section \ref{sec_finance}).

\begin{figure}
\centering
\includegraphics[scale=1]{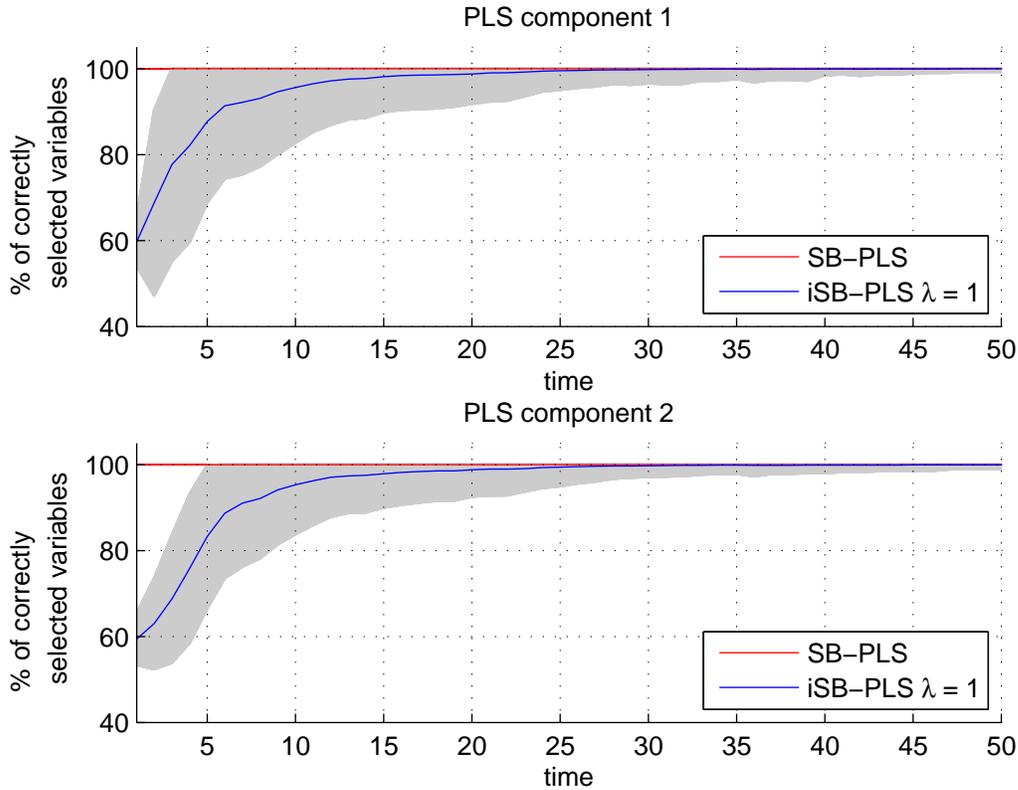}
\caption{Results of 500 runs with simulated static data showing the percentage of correctly estimated variables by SB-PLS (using the whole data set) and iSB-PLS (incrementally). The shaded area shows the Monte Carlo error (standard deviation) of correctly selected variables.}
\label{converge}
\end{figure}

Figure \ref{converge} shows the in-sample result of a Monte Carlo simulation consisting of 500 runs of the sparse Bridge PLS algorithm on simulated data with static coefficients. It can be seen that the off-line algorithm is able to correctly select all of the variables corresponding to the most important factor in both the first and second PLS components The blue line corresponds to the performance of the on-line iSB-PLS algorithm on the same data with a forgetting factor of 1. The shaded area shows the Monte Carlo error of the iSB-PLS result. It can be seen that the performance of the on-line algorithm quickly converges to the off-line algorithm within 35 data points. This suggests that after a brief learning period, the result obtained by the iSB-PLS algorithm is equivalent to that of the off-line algorithm in the case of stationary data, and they are both correct.

Furthermore, in order to test the adaptive behavior of the iSB-PLS algorithm using the input data streams described in Eq. \eqref{datastreams}, we generate an univariate output by introducing time-dependent regression coefficients. Until time $t=100$, all the variables associated with the first hidden factor strongly contribute to the output, and their regression coefficients are selected by sampling from a normal distribution centered at $10$ and with low variance. Analogously, the variables associated with the second hidden factors have regression coefficients with mean $5$ and with low variance. The variables associated with the third hidden factor are assigned zero coefficients. In order to introduce a non-stationary behavior, all the non-zero coefficients in the two groups of "active variables" are swapped at $t=101$. At $t=301$ until the end of the period, the first group of  variables is assigned a zero coefficient and the group associated with the third hidden factor is assigned a coefficient sampled from a normal distribution centred around 10.  In this way, the important predictors change over time and we expect these changes to be picked up in almost real-time by the algorithm. In this setting, we set $R=2$ and the sparsity parameter $\gamma$ is chosen automatically by the algorithm so that, at any given time, exactly $20$ variables are selected. The forgetting factor $\lambda$ is set to 0.98 to ensure a rapid adjustment when the coefficients switch while also keeping the switching frequency low to gain stability in the selected variables.
 
\begin{figure}
\centering
\epsfig{scale=1, file=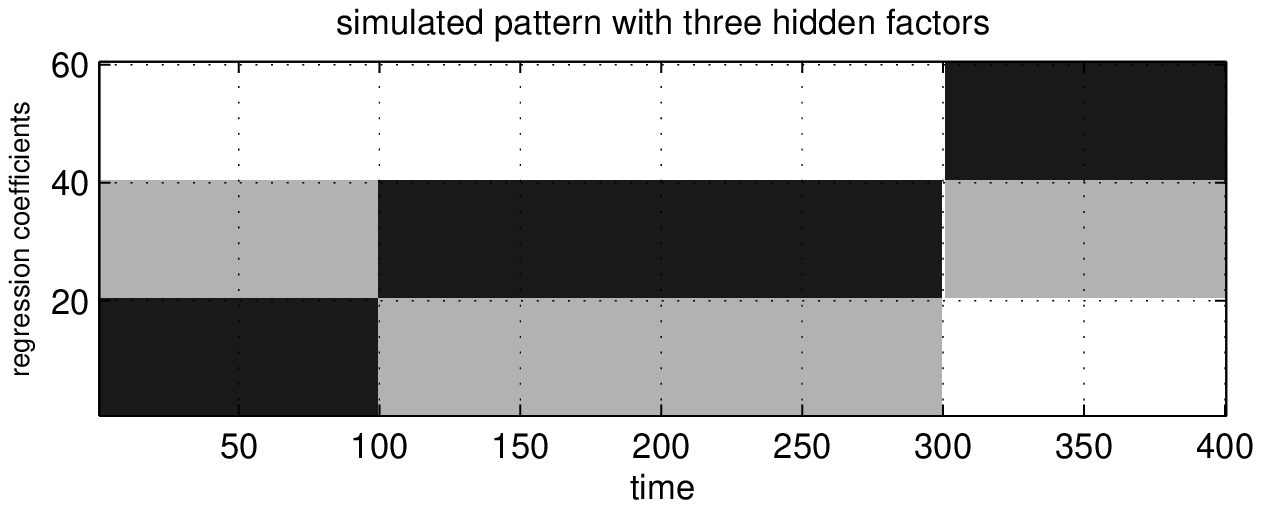}
\epsfig{scale=1, file=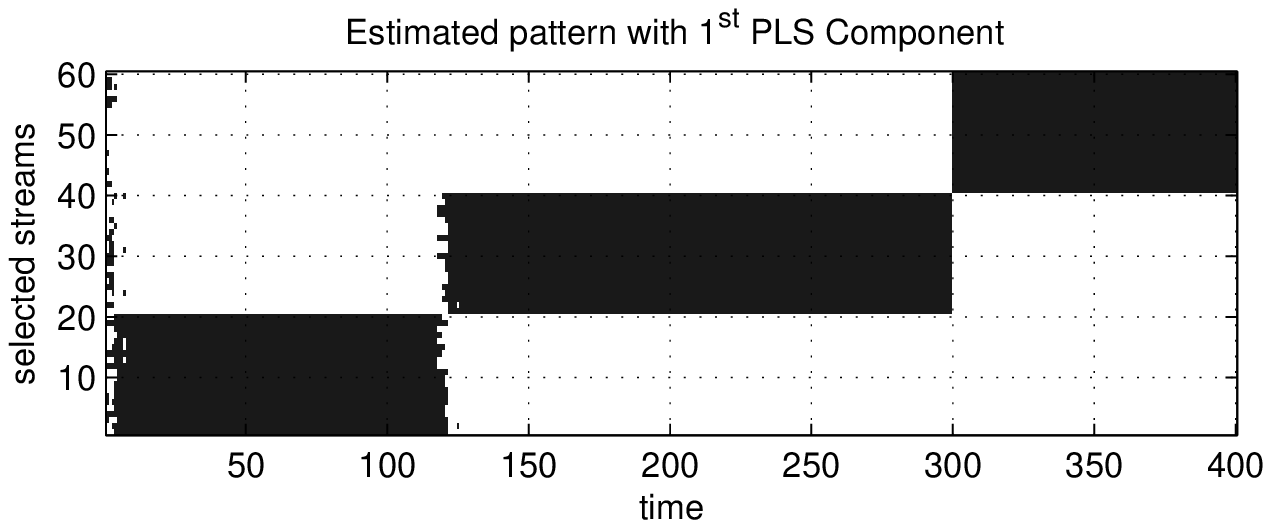}
\epsfig{scale=1, file=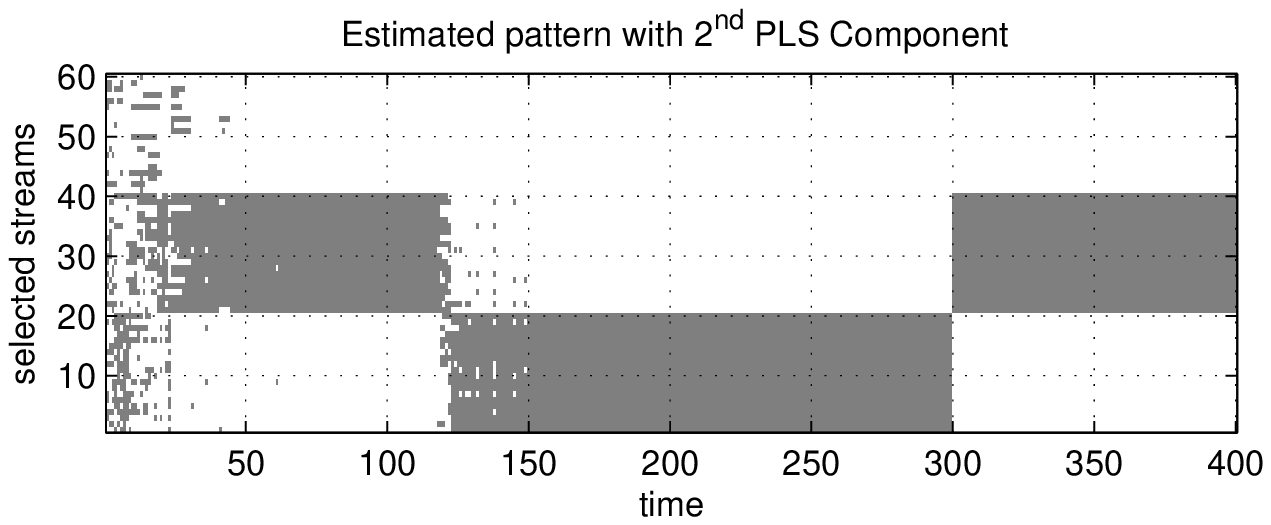}
\caption{Results of test with simulated data. The top figure shows
  how, at any time, there are three blocks of data streams: active
  streams having larger (black) and smaller (gray) regression
  coefficients, and inactive streams (white) which only contributes
  to noise. Each block is related to a different hidden factor. The
  bottom figure shows the data streams selected on-line by each PLS
  component.}
\label{simpleswitch}
\end{figure}

Figure \ref{simpleswitch} shows the results of a single run of this experiment. Clearly, the first PLS component is able to accurately select the most important group of variables. The second component always selects the second most important group of variables whilst mostly ignoring the group of variables selected by the first component. Neither component selects the inactive variables suggesting the algorithm is correctly able to distinguish important predictors from noise. As the coefficients switch, the algorithm only requires few data points before it detects the changes and adapts itself. Faster adaptation may be achieved by controlling the forgetting factor $\lambda$. 

\begin{figure}
\centering
\includegraphics[scale=1]{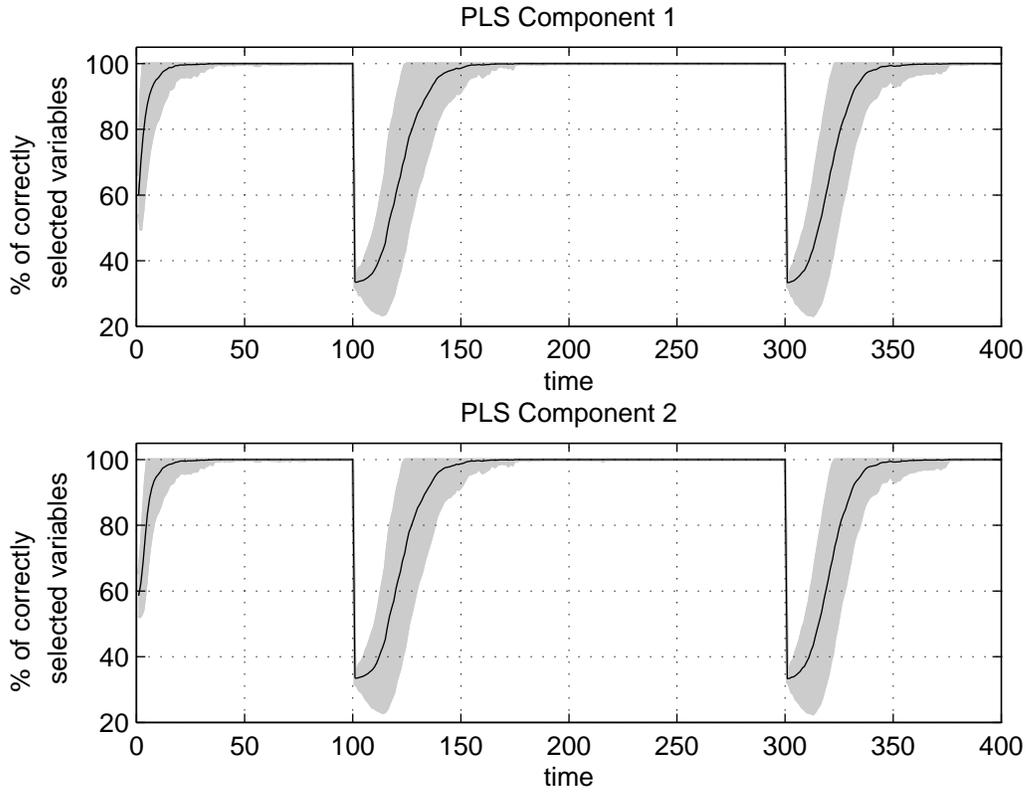}
\caption{Results of 500 runs with simulated data for $\lambda=0.98$. The solid line shows the mean percentage of correctly selected variables in each component. The shaded area shows the Monte Carlo error (standard deviation) of correctly selected variables.}
\label{percent_correct_pls}
\end{figure}

Figure \ref{percent_correct_pls} reports on the mean percentage of correctly selected variables in both components by the iSB-PLS algorithm in a Monte Carlo simulation consisting of 500 runs of this experiment. The solid line shows the mean percentage of correctly selected variables by the first and second PLS components. The shaded area shows the Monte Carlo error. It is clear that in the portions where the data is stationary, iSB-PLS will correctly select the important variables with very little error. In response to a change in the important factors, the percentage of correctly selected variables instantly decreases and quickly adapts to the new data. The algorithm eventually selects the correct variables after some settling time. However, during this time the variability of the result increases.

\begin{figure}
\centering
\includegraphics[scale=1]{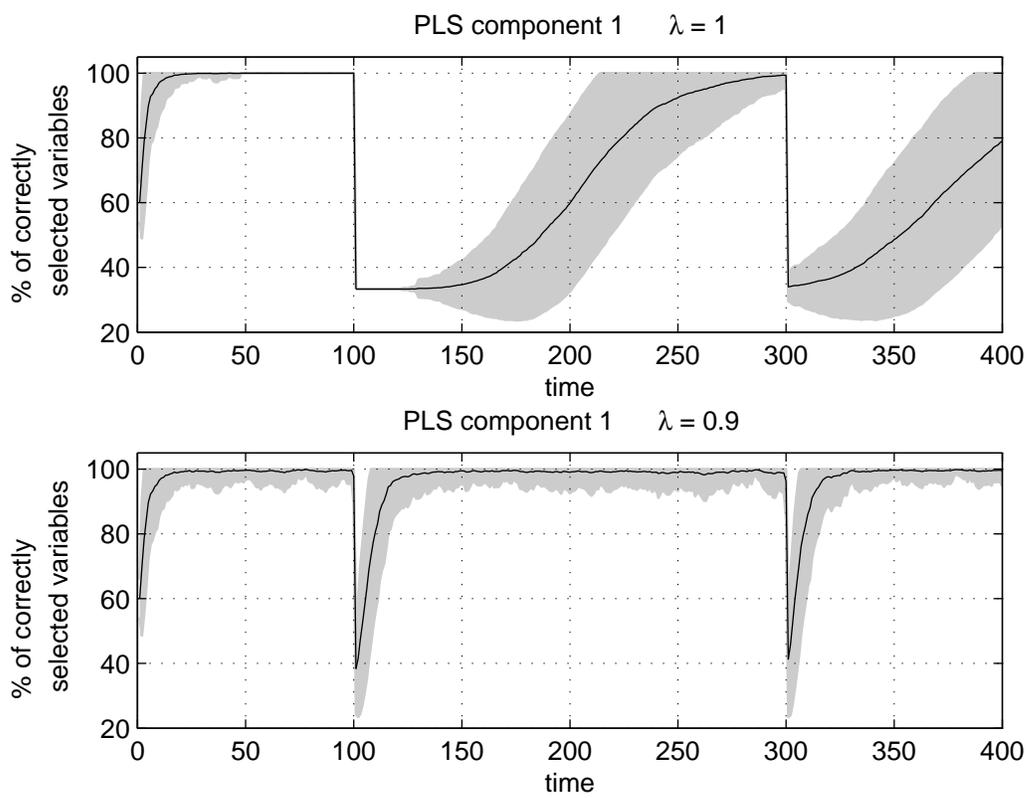}
\caption{Percentage of correctly selected variables by the first component for different values of $\lambda$.}
\label{different_lambda}
\end{figure}

Figure \ref{different_lambda} shows the effect of changing the forgetting factor, $\lambda$. When $\lambda=1$, no forgetting takes place and the algorithm is very slow to adapt to changes. When $\lambda=0.9$, the algorithm adapts to changes quickly. However a smaller forgetting factor causes the solution to become unstable as the algorithm is very sensitive to small changes and noise in the data. This can be seen by observing the larger Monte Carlo error during the periods of stationary data in the case where $\lambda=0.9$.

\subsection{An application to index tracking} \label{sec_finance}

An example application of the iSB-PLS algorithm lies in the financial domain and is related to the \emph{index tracking} problem. The objective of index tracking is to select a small portfolio of assets and determine weights, which represent a proportion of the total investment capital, so that the returns achieved by the portfolio track very closely those achieved by a benchmark index. Our application of a sparse algorithm to the portfolio selection and index tracking problem is supported by work in \cite{Brodie2008} who propose sparse portfolios based on Lasso penalized regression. Furthermore the use of a latent factor model for index tracking is supported by evidence which suggests that the first principal component of index returns captures the \emph{market factor} (see, for example  \cite{Alexander2005}). Our framework unifies these two approaches by combining dimensionality reduction by projection onto latent factors with variable selection using a regularized regression. For this application, we use published data from the S\&P and\ Nikkei indices as described in \cite{Beasley2003}.

\begin{figure}
\centering
\includegraphics[scale=1]{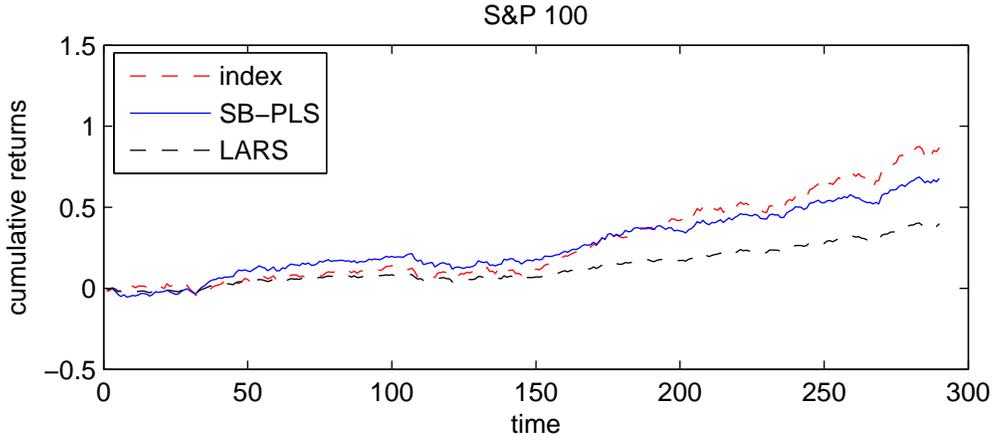}

\caption{Comparison of enhanced tracking (+15\% annual returns) of the S\&P using a static portfolio of 10 stocks chosen using SB-PLS and LARS.}

\label{lars}
\end{figure}

To motivate the use of an incremental algorithm for index tracking, we present an example of tracking with two off-line methods. We perform "enhanced tracking" (see, for instance, \cite{Alexander2008}) of the S\&P index. This consists of performing index tracking in the case where the target asset to be tracked are the index returns plus an additional $15\%$ annual returns. We use the LARS algorithm of \cite{LARS} and our sparse Bridge PLS algorithm with one latent factor.  Figure \ref{lars} shows the in-sample results of enhanced tracking of the S\&P100 index using a static portfolio of 10 stocks selected from 98. Despite using the in-sample result, it is clear that using a static portfolio for a long period of time leads to poor tracking performance and in both cases the artificial portfolios underperform the index. This is due to the financial index being non-stationary and suggests that a scheme for \emph{rebalancing} the portfolio would produce better tracking performance.  

We have tested the iSB-PLS algorithm in a more involved setting where: (a) two indices (the S\&P and the Nikkei) need to be simultaneously tracked, so the response is bivariate, and (b) both benchmark indices have been \emph{enhanced} as previously described. The total number of available stocks is $323$ and we set the portfolio size to $10$. The forgetting factor is $\lambda=0.99$ and we constrain the selected stock to be associated to the main latent factor only, so that $R=1$, as in \cite{Alexander2005}.

In order to assess whether our procedure selects and tracks the important variables over time, we compare its performance with the average returns obtained from a population of $1000$ portfolios of the same size, with each portfolio being made of a randomly selected subset of assets. To make sure that the comparison is fair, the portfolio weights are also time-varying and are obtained by using a recursive least squares method with the same $\lambda$ parameter. This comparison is made in order to determine whether the ability to update the portfolio composition in response to perceived changes in the market is really advantageous in an index tracking application.

\begin{figure}
\epsfig{scale=1, file=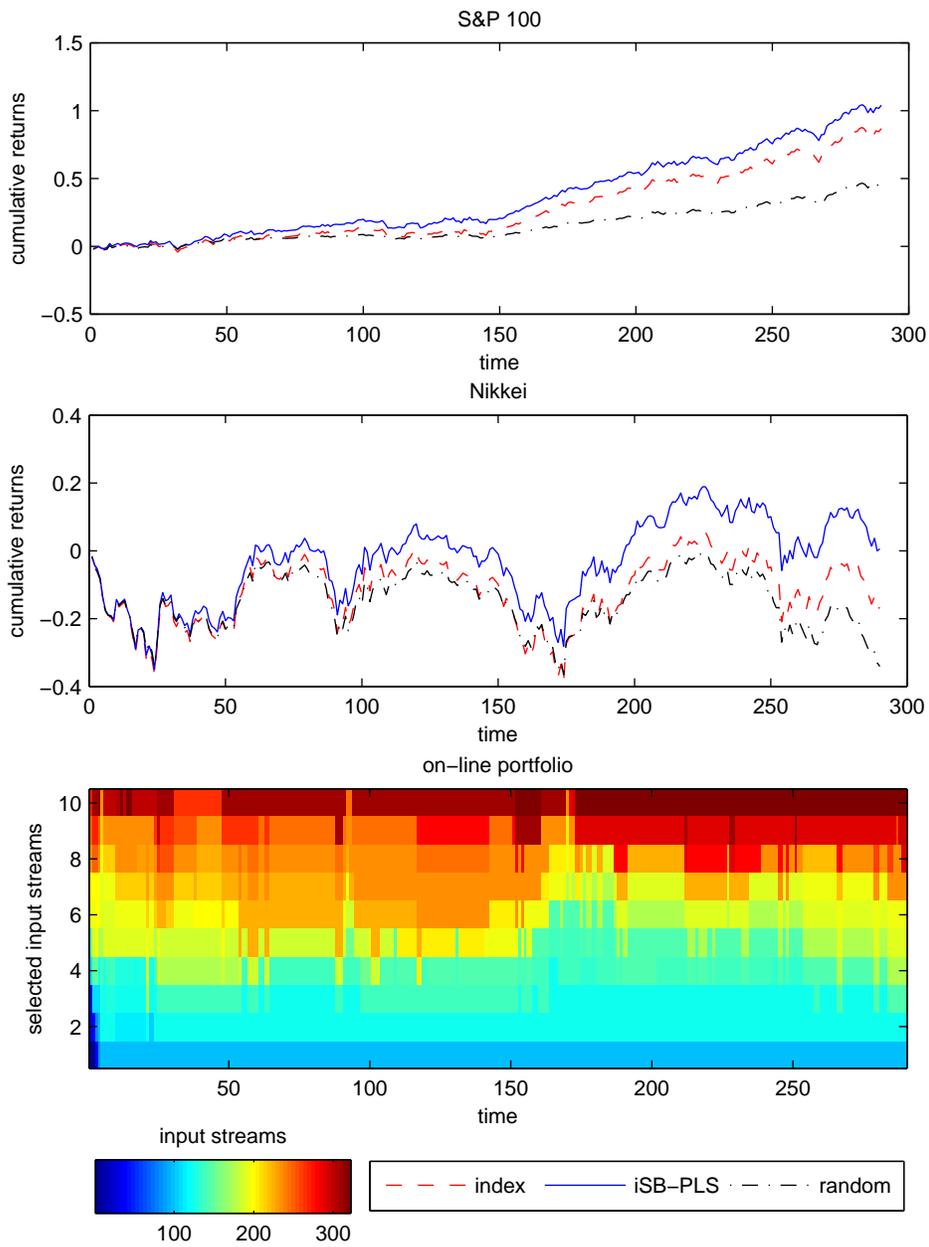}
\caption{Bivariate enhanced tracking  (+15\%
  annual returns) of the S\&P and Nikkei\ indices using a dynamic portfolio of $10$ stocks.}
\label{plus15}
\end{figure}

Figure \ref{plus15} shows the results of this test.  It can be seen that iSB-PLS consistently overperforms both indices and selects a small portfolio achieving exactly the target annual returns of $+15\%$. In comparison, the random portfolio underperforms the S\&P index by $32.07\%$ and the Nikkei by $8.42\%$. Our results suggest that the importance of certain stocks in the index is not constant over time so the ability to detect and adapt to these changes is certainly advantageous. Using a model that assumes a time-varying latent factor driving the asset returns is also advantageous in this
setting, since its existence in real markets has been heavily documented in the financial literature. The bottom plot of Figure \ref{plus15} is a heatmap illustrating how the make-up of the portfolio selected by iSB-PLS changes during the entire period. Specifically, it shows the existence of a few important stocks that are held for the majority of the period whereas other assets are picked and dropped more frequently throughout the period, further suggesting that it is advantageous to be able to adapt the constituents of a tracking portfolio. However, associated with every change made to the portfolio are transaction costs. If too many changes take place, the costs will outweigh the returns so an intelligent rebalancing strategy must be developed which finds a trade-off between good tracking and low transaction costs. 

\section{Conclusions} \label{conclusions}

In this work we have presented an on-line algorithm for variable selection in a multivariate regression context based on streaming data. As far as we are aware, this is the first such algorithm which combines dimensionality reduction and variable selection for data streams in a unified framework. From the simulation results we have shown that the algorithm is able to accurately select variables associated with the important factors underlying the data. In the case of non-stationary data where the important factors are changing, iSB-PLS is able to accurately track the changes. 


We have identified a number of open questions and avenues for further research. iSB-PLS requires the specification of a number of parameters which are currently pre-specified by the user. The question of how to select, in an on-line and adaptive manner, the number of PLS components and the number of variables per component is an important one and we are currently working towards the development of self-tuning procedures.

There are several methods in the literature for automatically updating the individual model parameters.  A mechanism for adapting the sparsity parameter, $\gamma_{t}$ at each time point was proposed by \cite{Anagnostopoulos2008}. They achieve this by evaluating the Akaike information criterion (AIC) of the model with a value of $\gamma_{t-1}$, $\gamma_{t-1} + c$ and $\gamma_{t-1} - c$, where $c$ is some small constant. The value of $\gamma_{t}$ which is used at that time point is the one which minimizes the AIC. This method could be incorporated into iSB-PLS as a simple adaptive solution to the variable selection portion of the model selection problem.

A method for selecting the number of principal components on-line using the concept of signal energy was proposed by \cite{Sun2006}. The energy at time $t$, $E_{t}$ is defined as the variance of the sequence up to $x_{t}$. The retained energy $\hat{E}_{t}$ is defined as the variance of the reconstructed sequence up to $x_{t}U_{t}$. The algorithm ensures that the retained energy is within the bounds $f_{E}E_{t}<\hat{E}_{t}<F_{E}E_{t}$. The upper and lower bounds are chosen so that retained energy is between 95\% and 98\% of the true energy of the signal. If the retained energy is too low, a new principal component is added to the model. Likewise, if the retained energy is too high, the least important principal component is removed from the model. A similar method for incremental PLS could be implemented for iSB-PLS.

A method to select the number of PLS projections on-line was proposed by \cite{Vijayakumar2005} who use an approximation of leave-one-out cross validation. The algorithm initially sets the number of projections, $R=2$ and recursively keeps track of a mean squared error term, $e_{t}^{(r)}$ as a function of the number of components, using a forgetting factor in the following way
\[
e_{t+1}^{(r)}=\lambda e_{t}^{(r)} + (y_{t}-\hat{y}_{t})^{2}
\]
where $\hat{y}_{t}$ is the estimated response at time, $t$. 
If at time $t+1$ adding a new PLS component causes a large enough reduction in error, the number of PLS components is increased. If adding the new component does not decrease the error enough, the number of PLS components is not changed.

Since both parameters must be selected and updated so that the correct number of factors and the correct number of variables per factor are chosen, there needs to be a unified framework for measuring the model fit and determining what parameters need to be changed and when. We have identified one potential way to achieve this by monitoring the percentage of explained covariance between $X$ and $Y$ at every time point. Since PLS maximizes the covariance between $X$ and $Y$, if the monitored percentage of explained covariance becomes lower than some threshold the model parameters should be updated. \cite{Shen2008} describe a method for quantifying the percentage of variance accounted for by sparse principal components. However, it remains to be seen whether this can be adapted for iSB-PLS.

The forgetting factor $\lambda$ has also been pre-selected, however a number of techniques exist for learning this parameter from the data in a streaming fashion. These techniques have been discussed in the literature concerning on-line learning of neural networks, as in \cite{Saad1999}, and  other time-varying processes, as in \cite{Niedzwiecki2000}. Furthermore, we are planning to apply these methods to related financial applications such as further extensions of index tracking for building \emph{market neutral} portfolios and detecting market inefficiencies for algorithmic trading, as in \cite{Alexander2008} and \cite{Montana2008a}, respectively. We are considering other applications in the field of text mining involving news feeds.

\bibliographystyle{abbrv}
\bibliography{plspreprint_v1}  

\begin{thebibliography}{10}

\bibitem{Alexander2005}
C.~Alexander and A.~Dimitriu.
\newblock Sources of over-performance in equity markets: mean reversion, common
  trends and herding.
\newblock Technical report, ISMA Center, University of Reading, UK, 2005.

\bibitem{Alexander2008}
C.~Alexander and A.~Dimitriu.
\newblock Equity indexing: Optimize your passive investments.
\newblock {\em Quantitative Finance}, 4(3), 2008.

\bibitem{Anagnostopoulos2008}
C.~Anagnostopoulos, D.~Tasoulis, D.~J. Hand, and N.~M. Adams.
\newblock Online optimisation for variable selection on data streams.
\newblock In {\em Proc. of the 18th European Conf. on Artificial Intelligence},
  2008.

\bibitem{Beasley2003}
J.~Beasley, N.~Meade, and T.~J. Chang.
\newblock An evolutionary heuristic for the index tracking problem.
\newblock {\em European Journal of Operational Research}, 148:621–643, 2003.

\bibitem{Brodie2008}
J.~Brodie, I.~Daubechies, C.~D. Mol, C.~Giannone, and I.~Loris.
\newblock Sparse and stable markowitz portfolios.
\newblock {\em European Central Bank Working Paper Series}, 936, 2008.

\bibitem{LARS}
B.~Efron, T.~Hastie, I.~Johnstone, and R.~Tibshirani.
\newblock Least angle regression.
\newblock {\em Annals of Statistics}, 32:407--499, 2004.

\bibitem{Erlich1994}
S.~Erlich and K.~Yao.
\newblock Convergences of adaptive block simultaneous iteration method for
  eigenstructure decomposition.
\newblock {\em Signal Processing}, 37, 1994.

\bibitem{Forsythe1976}
G.~Forsythe.
\newblock {\em Computer Methods for Mathematical Computations}.
\newblock Prentice Hall, 1976.

\bibitem{Friedman2007}
J.~Friedman, E.~Hastie, H.~H\"{o}fling, and R.~Tibshirani.
\newblock Pathwise coordinate optimization.
\newblock {\em The Annals of Applied Statistics}, 1(2):302--332, 2007.

\bibitem{Gidskehaug2004}
L.~Gidskehaug, H.~Stødkilde-Jørgensen, M.~Martens, and H.~Martens.
\newblock Bridge-{PLS} regression: two-block bilinear regression without
  deflation.
\newblock {\em Journal of Chemometrics}, 18, 2004.

\bibitem{Haykin2001}
S.~Haykin.
\newblock {\em Adaptive Filter Theory}.
\newblock Prentice Hall, 2001.

\bibitem{Hoskuldsson1988}
A.~Hoskuldsson.
\newblock Pls regression methods.
\newblock {\em Journal of Chemmometrics}, 2, 1988.

\bibitem{Kim2004}
S.-P. Kim, Y.~N. Rao, D.~Edogmus, and J.~C. Principe.
\newblock Tracking of multivariate time-variant systems based on on-line
  variable selection.
\newblock {\em 2004 IEEE Workshop on Machine Learning for Signal Processing},
  2004.

\bibitem{Le2008}
K.~L\^{e}~Cao, D.~Rossouw, C.~Robert-Grani\'{e}, and P.~Besse.
\newblock Sparse {PLS}: variable selection when integrating omic data.
\newblock Technical report, INRA, 2008.

\bibitem{Montana2008a}
G.~Montana, K.~Triantafyllopoulos, and T.~Tsagaris.
\newblock Data stream mining for market-neutral algorithmic trading.
\newblock In {\em Proceedings of the ACM Symposium on Applied Computing}, pages
  966--970, 2008.

\bibitem{Niedzwiecki2000}
M.~Nied\'{z}wiecki.
\newblock {\em Identification of time-varying processes}.
\newblock Wiley, 2000.

\bibitem{Papadimitriou2005}
S.~Papadimitriou, J.~Sun, and C.~Faloutsos.
\newblock Streaming pattern discovery in multiple time-series.
\newblock In {\em Proceedings of the 31st International Conference on Very
  Large Data Bases}, pages 697 -- 708, 2005.

\bibitem{Rosipal2006}
R.~Rosipal and N.~Kr\"{a}mer.
\newblock Overview and recent advances in partial least squares.
\newblock pages 34--51. 2006.

\bibitem{Saad1999}
D.~Saad, editor.
\newblock {\em On-Line Learning in Neural Networks}.
\newblock Number~17 in Publications of the Newton Institute. Cambridge, 1999.

\bibitem{Shen2008}
H.~Shen and J.~Huang.
\newblock Sparse principal component analysis via regularized low rank matrix
  approximation.
\newblock {\em Journal of Multivariate Analysis}, 2008.

\bibitem{Sun2006}
J.~Sun, S.~Papadimitriou, and C.~Faloutsos.
\newblock Distributed pattern discovery in multiple streams.
\newblock In {\em Proceedings of the Pacific-Asia Conference on Knowledge
  Discovery and Data Mining, Singapore}, 2006.

\bibitem{Vijayakumar2005}
S.~Vijayakumar, A.~D'Souza, and S.~Schaal.
\newblock Incremental online learning in high dimensions.
\newblock {\em Neural Computation}, 17:2602--2634, 2005.

\bibitem{Weng2003}
J.~Weng, Y.~Zhang, and W.~S. Hwang.
\newblock Candid covariance-free incremental principal component analysis.
\newblock {\em IEEE Transactions on Pattern Analysis and Machine Intelligence},
  25(8):1034--1040, 2003.

\end{thebibliography}

\end{document}